\documentclass{article}
\usepackage{spconf,amsmath}
\usepackage{graphicx}
\usepackage{url}
\usepackage{amsmath}
\usepackage{amssymb}
\usepackage{amsfonts}
\usepackage{latexsym}
\usepackage{enumerate}
\usepackage{mathtools}
\usepackage{booktabs}
\usepackage{multirow}
\usepackage{cite} 
\usepackage{arydshln}
\usepackage{tablefootnote}
\usepackage{threeparttable}
\usepackage[hang]{footmisc}

\title{Leveraging Large Text Corpora for End-to-End Speech Summarization}

\name{\begin{tabular}{c}
  Kohei Matsuura, Takanori Ashihara, Takafumi Moriya, Tomohiro Tanaka, \\Atsunori Ogawa, Marc Delcroix, Ryo Masumura
\end{tabular}}
\address{NTT Corporation, Japan}
\begin{document}
\ninept

\maketitle

\begin{abstract}
End-to-end speech summarization (E2E SSum) is a technique to directly generate summary sentences from speech.
Compared with the cascade approach, which combines automatic speech recognition (ASR) and text summarization models, the E2E approach is more promising because it mitigates ASR errors, incorporates nonverbal information, and simplifies the overall system.
However, since collecting a large amount of paired data (i.e., speech and summary) is difficult, the training data is usually insufficient to train a robust E2E SSum system.
In this paper, we present two novel methods that leverage a large amount of external text summarization data for E2E SSum training. 
The first technique is to utilize a text-to-speech (TTS) system to generate synthesized speech, which is used for E2E SSum training with the text summary.
The second is a TTS-free method that directly inputs phoneme sequence instead of synthesized speech to the E2E SSum model.
Experiments show that our proposed TTS- and phoneme-based methods improve several metrics on the How2 dataset.
In particular, our best system outperforms a previous state-of-the-art one by a large margin (i.e., METEOR score improvements of more than 6 points).
To the best of our knowledge, this is the first work to use external language resources for E2E SSum. 
Moreover, we report a detailed analysis of the How2 dataset to confirm the validity of our proposed E2E SSum system.

\end{abstract}

\vspace{-3pt}
\section{Introduction}
\label{sec:intro}
\vspace{-3pt}
With the introduction of deep learning, automatic speech recognition (ASR) technology has made significant progress \cite{asr1}.
Despite its success, ASR transcriptions are not highly readable because they often contain fillers, disfluencies, or redundant expressions \cite{goldman2005}.
To generate more readable and informative text, speech summarization (SSum) technology has attracted much attention for meetings \cite{murray2010}, educational videos \cite{Sharma2021}, and patient-physician conversations \cite{finley2018}.

\begin{figure}[t]
  \vspace{5pt}
  \centering
  \includegraphics[scale=0.49]{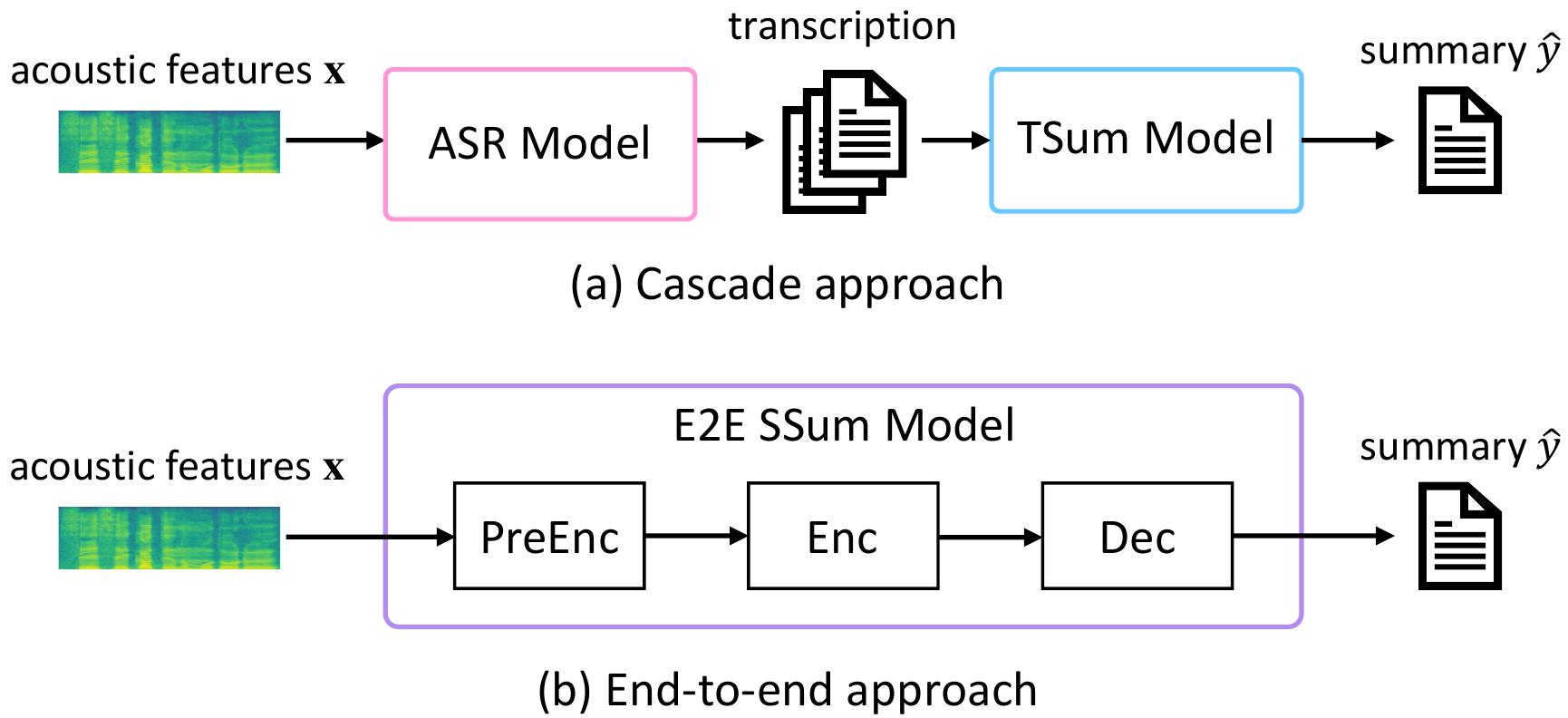}
  \caption{Comparison between the (a) Cascade and (b) End-to-end (E2E) approaches for speech summarization.}
  \label{fig:cascade_vs_e2e}
  \vspace{-5pt}
\end{figure}

In general, SSum has been achieved with the cascade approach, which combines ASR and text summarization (TSum) models, as shown in Fig. \ref{fig:cascade_vs_e2e}(a). 
Cascade SSum achieves high performances with a combined system of highly-accurate ASR models and TSum models pre-trained on large amounts of unpaired text data \cite{zhang2021,Zhong2021}.
Despite the success of the cascade approach, it suffers from ASR error propagation and a lack of nonverbal and acoustical information \cite{Tundik2019}.
To tackle these problems, \cite{Kano2021} fed a TSum model with N-best ASR hypotheses to mitigate the effect of ASR errors, and \cite{Liu2019} added acoustic features to the input text.

More recently, end-to-end SSum (E2E SSum) has also been proposed to address the aforementioned problems.
The E2E SSum system is similar to the E2E ASR system, which jointly optimizes acoustic and language models, and generates an abstractive summary directly with a single model as illustrated in Fig. \ref{fig:cascade_vs_e2e}(b).
Since this method does not include text as an intermediate output, the model can utilize fully acoustic information to generate summaries and is probably free from ASR errors.
\cite{Sharma2022} used this approach with restricted self-attention and reported that it performed better than the cascade approach.

However, the E2E SSum model requires a large number of costly speech-summary pairs, while most speech summarization tasks have a limited amount of pair data.
We assume that the huge external language resources could improve the E2E SSum models as in other E2E speech processing fields \cite{Mimura2018,Ueno2021,Hayashi2018,Ueno2021_2,Renduchintala2018,Masumura2020,jia2019}. 

In this paper, we attempt to leverage TSum data, i.e., Gigaword \cite{graff2003,Rush2015}, to enable the model to learn better linguistic representation.
We propose a text-to-speech (TTS)-based speech augmentation approach that synthesizes speech from the text of an external TSum dataset to obtain pseudo-speech and summary paired data for E2E SSum training.
Although the application of this technique is rather simple, synthesizing pseudo-speech requires high computational costs.
To eliminate these costs, we also propose a method that extends the model with a phoneme pre-encoder, inspired by \cite{Renduchintala2018,Masumura2020}. 
The phoneme pre-encoder is separated from the speech pre-encoder when we input phoneme sequences instead of synthesized speech.
This method requires much less computational cost and storage space than TTS-based data augmentation.

All experiments are conducted on the How2 dataset \cite{Sanabria2018} composed of short video clips and their summaries.
We performed a deep analysis of the corpus and found that part of the training and evaluation data were very similar, which could bias the results. 
We thus propose an evaluation scheme that filters out such data using quantitative metrics. 
With the proposed evaluation, we can attest summarization performance with more confidence.

Our major contributions are listed as follows:
\begin{enumerate}
  \item{Our best system achieves state-of-the-art performance on the How2 summarization task.}
  \item{We confirm that external TSum data enables E2E SSum model improvements with TTS technology.}
  \item{We propose a phoneme-based data augmentation method, resulting in lower computational costs than TTS-based one.}
  \item{We suggest that filtering the evaluation set of the How2 dataset is important for assessing the quality of summarization abilities of models more reasonably.}
\end{enumerate}

\vspace{-3pt}
\section{Related work}
\vspace{-3pt}
\label{sec:format}
Previous studies have demonstrated the effectiveness of utilizing external text data for E2E ASR and E2E speech translation systems. 
For E2E ASR systems, \cite{Mimura2018} generated additional speech-transcription data by using Tacotron2 \cite{Shen2018} to synthesize acoustic features from external text data in the context of domain adaptation. 
Other studies utilized text data via discrete speech representations extracted with vq-wav2veq \cite{Ueno2021}, latent representations of E2E ASR encoders \cite{Hayashi2018}, and phoneme sequences \cite{Renduchintala2018,Masumura2020}.
Furthermore, \cite{Tjandra2019} proposed a method called ``Speech Chain'' that jointly optimizes TTS and ASR models with additional unpaired text and speech. 
\cite{Wang2020} expanded this method with a consistency loss.
Similar ideas were also used to generate training data for E2E speech translation systems \cite{jia2019}.
Our study is inspired by these successes and attempts to apply data augmentation approaches to an E2E SSum system.

\vspace{-3pt}
\section{E2E SSum Model}
\vspace{-3pt}
\label{sec:models}
As shown in Fig. \ref{fig:cascade_vs_e2e}, the E2E SSum system unifies the ASR and TSum models into a single model, while the cascade system processes input speech via transcriptions.
Specifically, we adopt the transformer-based attentional encoder-decoder (AED) model \cite{asr1} in this paper.
First, the pre-encoder $\mathrm{PreEnc}$ embeds acoustic features $\mathbf{x}$ into a sub-sampled sequence of hidden representations $\mathbf{h}^{\mathrm{pre}}$:
\begin{eqnarray}
  \mathbf{h}^{\mathrm{pre}} = \mathrm{PreEnc}(\mathbf{x}),
\end{eqnarray}
where $\mathrm{PreEnc(\cdot)}$ is composed of 2-D convolutional neural networks (CNNs).
Second, the encoder $\mathrm{Enc}(\cdot)$ calculates a latent representation $\mathbf{h}^{\mathrm{enc}}$ from $\mathbf{h}^{\mathrm{pre}}$:
\begin{eqnarray}
  \mathbf{h}^{\mathrm{enc}} = \mathrm{Enc}(\mathbf{h}^{\mathrm{pre}}, \mathbf{R}),
\end{eqnarray}
where $\mathrm{Enc}(\cdot)$ consists of Conformer \cite{gulati2020} blocks that achieve high accuracy in the ASR task, and $\mathbf{R}$ denotes the relative positional embeddings.
Finally, the decoder $\mathrm{Dec}(\cdot)$ receives $\mathbf{h}^{\mathrm{enc}}$ and previous outputs $\{\hat{y}_i\}_{i=1}^{l-1}$ as a clue to autoregressively infer each token:
\begin{eqnarray}
  \mathbf{h}^{\mathrm{dec}}_l = \mathrm{Dec}([\mathrm{Emb}(\hat{y}_i) + \mathrm{PE}_i]_{0 \le i \le l-1}, \mathbf{h}^{\mathrm{enc}}),
\end{eqnarray}
\vspace{-20pt}
\begin{eqnarray}
  \hat{\mathbf{y}}_l = \mathrm{softmax} (\mathbf{h}^{\mathrm{dec}}_{l}),
\end{eqnarray}
where $\mathrm{softmax}(\cdot)$ is the softmax activation function, and $\mathrm{PE}_i$ is the $i$-th absolute positional encoding \cite{Vaswani2017}.
The decoder $\mathrm{Dec}(\cdot)$ is composed of Transformer blocks, and the token embedding module $\mathrm{Emb}(\cdot)$ is a learnable linear layer.
The whole summary sentence $\hat{y}$ is estimated by a beam search technique.
At the beginning of the inference, the decoder receives the special symbol \texttt{<sos>}, which indicates the start of the sentence, and it outputs tokens until the special end token \texttt{<eos>} is estimated.
During training, the E2E SSum model is optimized by the cross-entropy loss between $\hat{\mathbf{y}}_l$ and the one-hot vector of $y_l$.
To stabilize model training, the ground truth token $y_i$ is used in Eq. (3) instead of $\hat{y}_i$.

In addition to the aforementioned architecture, we apply the following two modifications to the E2E SSum model.
First, we replace batch normalization (BN) in the Conformer blocks with layer normalization (LN).
Since the input lengths of the SSum system are very long, the batch size is too small to learn well-generalized statistics of the BN layers during training.
Second, we propose to replace the absolute positional encoding in Eq. (3) with the learnable positional embedding (LPE) \cite{Gehring2017} as
\begin{eqnarray}
  \mathbf{h}^{\mathrm{dec}}_l = \mathrm{Dec}([\mathrm{Emb}(\hat{y}_i) + \mathrm{LPE}(i)]_{0 \le i \le l-1}, \mathbf{h}^{\mathrm{enc}}),
\end{eqnarray}
where a learnable linear layer $\mathrm{LPE}(i)$ embeds the position of the $i$-th token.
Since the LPE has been successfully used in the natural language processing field \cite{Gehring2017,lewis2020}, we expect LPE could also improve performance for the SSum task.

\begin{figure}[t]
  \centering
  \includegraphics[scale=0.47]{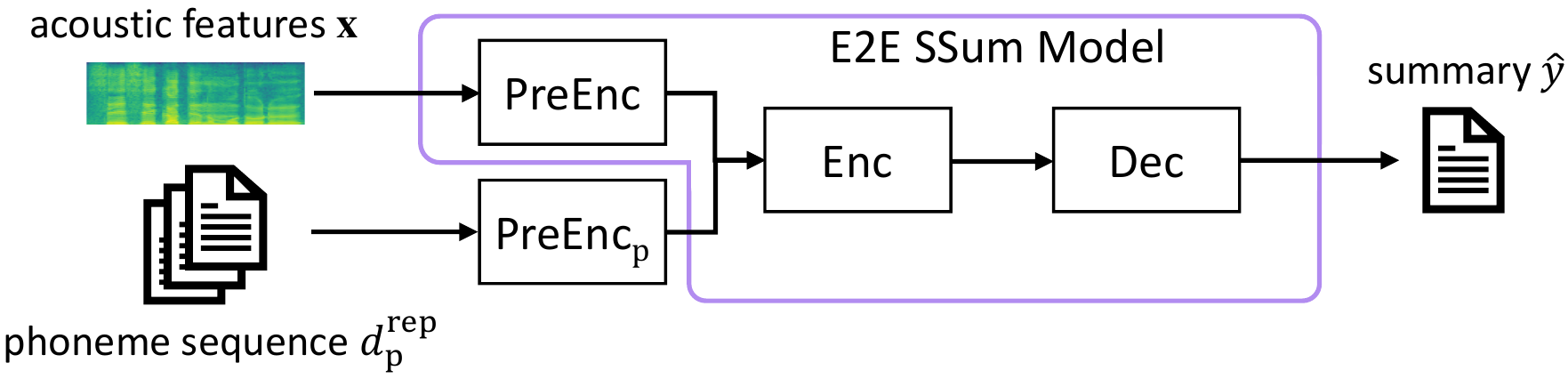}
  \caption{E2E SSum model with phoneme pre-encoder $\mathrm{PreEnc}_{\mathrm{p}}$.\vspace{-10pt}}
  \label{fig:preenc_p}
  \vspace{-10pt}
\end{figure}

\begin{table*}[ht!]
  \caption{Results of previous work (P-1) \cite{Sharma2022}, our baseline (C-1, B-1, B-2), and proposed models (T-1, 2).}
  \centering
  \begin{tabular}{c|c|ccc|ccc}
    \hline
      &ID & Model &\begin{tabular}{c} augmented data \end{tabular} & \#params. & \begin{tabular}{c}ROUGE-1, 2, L \end{tabular}($\uparrow$)& METEOR ($\uparrow$) & BERTScore ($\uparrow$)\\
    \hline\hline
    \multirow{4}{*}{baseline}  &C-1 & cascade &\multirow{4}{*}{-}&201M+140M & 60.9, 43.3, 55.4 & 30.3 & 92.6\\
      &P-1 & \cite{Sharma2022}~ & &104M & 60.9, 43.0, 55.9 & 28.8 & 91.0\\
      &B-1 & base & &98M & 65.3, 51.4, 62.1 & 32.5 & 93.0\\
      &B-2 & large & &203M & 65.6, 50.9, 62.0 & 32.8 & 93.2\\
    \hdashline[3pt/2pt]
    \multirow{2}{*}{proposed}&T-1 & large & TTS &203M & \textbf{68.4}, \textbf{54.1}, \textbf{65.0} & \textbf{34.9} & \textbf{93.8}\\
      &T-2 & large & phoneme &203M(+7M)\footnotemark & 67.4, 53.2, 64.1 & 33.9 & 93.6\\
    \hline
  \end{tabular}
  \label{table:results}
  \vspace{-10pt}
\end{table*}

\vspace{-3pt}
\section{Proposed Methods}
\label{sec:methods}
\vspace{-3pt}
\subsection{Multi-stage training}
\label{sec:multi}
\vspace{-3pt}
Our E2E SSum model receives three training datasets: paired data of speech and transcription $(\mathbf{x}, y_{\mathrm{asr}}) \in \mathcal{R}_\mathrm{ASR}$, paired data of speech and summary $(\mathbf{x}, y) \in \mathcal{R}_\mathrm{Sum}$, and external text summarization data $(d, y) \in \mathcal{R}_\mathrm{Ext}$. 
Here, $d$ is a TSum input text.
To leverage a variety of knowledge, we adopt a training strategy consisting of multiple stages as follows:
\begin{enumerate}[(i)]
  \item An AED model is trained as an ASR model using the pairs in $\mathcal{R}_\mathrm{ASR}$ to acquire the transcribing capability internally.
  \item To build the E2E SSum model, the aforementioned model is additionally fine-tuned with $\mathcal{R}_\mathrm{Sum}$.
  \item The model is further fine-tuned with the combined dataset of $\mathcal{R}_\mathrm{Sum}$ and artificial data derived from $\mathcal{R}_\mathrm{Ext}$.
\end{enumerate}
We define the model after stage (ii) as a baseline model and after stage (iii) as an augmented model by the TTS- or phoneme-based method.

\vspace{-3pt}
\subsection{TTS- and phoneme-based data augmentation}
\vspace{-3pt}
To utilize the TSum data for E2E SS, we propose two approaches: TTS- and phoneme-based data augmentation.
For the TTS-based approach, we synthesize the raw waveforms from the input texts $d$ of the TSum dataset and extract acoustic features for the stage (iii).

Although this TTS-based method is straightforward to augment speech summarization data, it requires high computational costs to synthesize speech.
Therefore, we also propose an alternative approach that feeds phoneme sequences into the E2E SSum model instead.
This approach is motivated by the performance improvement in \cite{Renduchintala2018}, in which a multi-modal (i.e., acoustic and textual) data augmentation technique for E2E ASR was proposed.
As a procedure, a word sequence $d$ is first automatically converted into a phoneme sequence $d_p$ by utilizing a grapheme-to-phoneme model.
The phoneme duration is then modeled by repeating each phoneme in $d_p$ in accordance with separately calculated statistics.
The repeated phoneme sequence is denoted by $d^{\mathrm{rep}}_p$.
To input the $d^{\mathrm{rep}}_p$ together with $\mathbf{x}$, we also add and optimize the phoneme pre-encoder $\mathrm{PreEnc}_{\mathrm{p}}(\cdot)$ in addition to $\mathrm{PreEnc}$ simultaneously as
\begin{eqnarray}
  \vspace{-3pt}
  \mathbf{h}^{\mathrm{pre}} = 
  \begin{cases}
    \mathrm{PreEnc}(\mathbf{x})\\
    \mathrm{PreEnc}_{\mathrm{p}}(d^{\mathrm{rep}}_p).
  \end{cases}
  \vspace{-3pt}
\end{eqnarray}
Both inputs $\mathbf{x}$ and $d^{\mathrm{rep}}_p$ share other modules as depicted in Fig. \ref{fig:preenc_p}.
Here, we choose bidirectional LSTMs followed by a LN layer and subsampling CNNs for $\mathrm{PreEnc}_{\mathrm{p}}$. 

\vspace{-3pt}
\section{Experiments}
\vspace{-3pt}\vspace{-3pt}
\subsection{Datasets}
\label{sec:dataset}
\vspace{-3pt}
We used the How2 dataset, which has approximately 2,000 hours of speech and is suitable for the E2E SSum task because of its relatively short speech inputs and abstractive summaries with a high compression rate. 
The training, validation, and evaluation sets contain 68,336, 1,600, and 2,127 pairs, and the average durations of their spoken documents are 84.7, 76.0, and 98.7 seconds, respectively.
The input speech is truncated up to 100 seconds and composed of 40-dimensional log Mel-filterbank outputs with 3-dimensional pitch features.
According to \cite{Sanabria2018}, the summaries are automatically collected from ``description" tags on YouTube.

For the data augmentation, we utilized the Gigaword corpus \cite{graff2003,Rush2015} composed of 3.8M pairs of the first sentences and titles of news articles.
The average word counts on this dataset for the input text and summaries are 31.4 and 8.3, respectively.
Since the average input length is much smaller than that of other text summarization corpora, such as CNN Daily Mail \cite{Hermann2015}, the synthesizing cost for the Gigaword corpus is relatively lower.
Even though the domain of this external corpus mismatches that of the How2 dataset, we expect that the model could learn additional words and expressions.

\footnotetext{The phoneme encoder (7M) is not needed during inference.}

\vspace{-3pt}
\subsection{Training and evaluation}
\vspace{-3pt}
We adopted a 12-layer Conformer encoder and a 6-layer Transformer decoder for the model architecture.
The pre-encoder was composed of 4-layer CNNs with a subsampling rate of 4.
We evaluated two model sizes: the base and large models with 512- and 768-dimensional embeddings, respectively.
Their encoders commonly had 2048 feed-forward (FF) units, 8 attention heads, and a kernel size of 31.
The decoder of the base model had 2048 FF units and 4 attention heads, and that of the large model had 3072 FF units and 12 attention heads.

In the training stage (i) described in Section \ref{sec:multi}, we trained the E2E ASR model with the Adam optimizer \cite{Kingma2015} and Noam scheduler \cite{Vaswani2017} with a learning rate of $2 \cdot 10^{-3}$, warmup steps of 40k, a weight decay rate of $10^{-5}$, and a batch size of 512.
The word error rates ($\downarrow$) of the base and large models on the evaluation set were 9.8\% and 9.5\% with a beam size of 16, respectively.

For the training stage (ii), the model was fine-tuned as the E2E SSum model with a learning rate of $10^{-4}$ with a reduction factor of 0.5, and batch size of 30.

The model was further fine-tuned with augmented data in the training stage (iii).
For the TTS model, we used VITS \cite{kim2021} trained with the LJSpeech corpus \footnote{\url{https://github.com/espnet/espnet/tree/master/}\\\url{egs2/ljspeech/tts1#pretrained-models-1}}.
The synthesized speech does not have a diversity of speakers because the LJSpeech contains only a single speaker.
This enables us to evaluate only the effectiveness of additional linguistic information excluding the effect of augmenting speaker varieties.
After generating the raw waveform, we extracted 43-dimensional acoustic features with a window size of 25ms and a shift of 10ms from the waveform and applied cepstral mean-variance normalization (CMVN) to match the original features in the How2 corpus.
When training the model in the stage (iii), we adopted a learning rate of $10^{-4}$ with a reduction factor of 0.5, and the learning rate of its decoder was $10^{-3}$. 
The maximum total length of input sequences in one batch was set to 300,000.
Each batch contains only real or artificial samples.

For the phoneme-based method, we used the g2pE toolkit\footnote{\url{https://github.com/Kyubyong/g2p}} to convert word sequences into phoneme sequences.
The number of phoneme units was 42.
The duration of phoneme $p$ was determined following the normal distribution $\mathcal{N}(\mu_p, \sigma_p)$ at each time it appeared.
Here, the mean $\mu_p$ and standard deviation $\sigma_p$ of $p$ were estimated with the TIMIT corpus \cite{Lamel1989}.
The phoneme pre-encoder had 3-layer bidirectional LSTMs followed by LN and 4-layer CNNs with a subsampling rate of 4.
The learning rate of the phoneme pre-encoder was $10^{-3}$.
The other configurations were the same as those of the TTS-based method.
In the stage (i), (ii), and (iii), we used byte-pair encoding and set the vocabulary size to 1,000.

When evaluating model performance, we chose the checkpoints with the best validation accuracy and decoded them with a beam width of 4 for all conditions. 
For evaluation metrics, we selected the ROUGE \cite{lin2004}, METEOR \cite{denkowski2014}, and BERTScore \cite{Zhang2020} scores commonly used in TSum tasks.

We also evaluated the conventional cascade model.
The ASR model was the same as the large model obtained after the training stage (i).
The TSum model was a BART-based model fine-tuned with the CNN/Daily Mail corpus\footnote{\url{https://huggingface.co/ainize/bart-base-cnn}} \cite{Hermann2015}
 and the How2 dataset. 
During the fine-tuning with the How2 dataset, we trained the model for 1M steps with the Adam optimizer, a learning rate of $5\times10^{-5}$ with linear decay, and a batch size of 8.
We used ESPnet2\footnote{\url{https://github.com/espnet/espnet}} for all of the implementation, training, and evaluation.

\vspace{-3pt}
\subsection{Results on entire How2 evaluation set}
\vspace{-3pt}
\label{sec:result1}
Table \ref{table:results} shows the evaluation scores of the augmented model (T-1, T-2) with prior state-of-the-art (P-1) \cite{Sharma2022}, the cascade model (C-1), and our baselines (B-1, B-2) on the How2 evaluation set.
Compared with Model P-1, Model B-1 with full attention, LN, and LPE remarkably improved the scores and achieved the current SoTA performance.
There was a slight improvement by enlarging the model size as in Model B-2, and this strong model is the baseline for the proposed data augmentation methods.
The result of Model T-1 shows that the TTS-based data augmentation method further improved the METEOR, ROUGE-L, and BERTScore scores by as much as 2.1, 3.0, and 0.6 points, respectively.
In addition, despite its low cost, Model T-2 trained with the phoneme-based method was also improved by 1.1, 2.1, and 0.4 points in terms of the METEOR, ROUGE-L, and BERTScore scores, respectively.
Although the domain of the augmented data is completely different from that of the How2 dataset, it helped the model expand its vocabulary.
For instance, as seen in Table \ref{table:improvement}, Model B-2 could not output a word \textit{gladiator}, which does not appear in the How2 training set, while Model T-1 and T-2 seemed to have learned it from the Gigaword training set.

\begin{table}[t]
  \vspace{-5pt}
  \caption{Part of summaries of ID: zpoDukAvPPY in the evaluation set generated by Model B-2, T-1, and T-2 with the ground truth (g.t.). The How2 training set does not contain the bolded word \textit{\textbf{gladiator}}, while the Gigaword training set contains it.}
  \begin{tabular}{cl}
    \hline
    \multirow{1}{*}{ID} & ~~~~~~~~~~~~~~~~~~~~~~~~~~~~~~~~~~~~~~Summary \\
    \hline\hline
    \multirow{2}{*}{B-2} & \textit{\small{learn to draw calf shoes with tips from a fashion expert}}\vspace{-1pt}\\
     & \textit{\small{in this free fashion design video.}}\vspace{3pt}\\
    \multirow{2}{*}{T-1} & \textit{\small{learn to draw \textbf{gladiator} sandals in this free fashion}}\vspace{-1pt}\\
     & \textit{\small{illustration video from a fashion design graduate student.}}\vspace{3pt}\\
    \multirow{2}{*}{T-2} & \textit{\small{learn to draw \textbf{gladiator} sandals with tips}}\vspace{-1pt}\\
     & \textit{\small{from a fashion expert in this free fashion design video.}}\vspace{3pt}\\
    \multirow{2}{*}{g.t.} & \textit{\small{learn to draw \textbf{gladiator} sandals in this free fashion video}}\vspace{-1pt}\\
     & \textit{\small{from a fashion design graduate student.}}\\
    \hline
  \end{tabular}
  \vspace{-5pt}
  \label{table:improvement}
\end{table}

\vspace{-3pt}
\subsection{Analysis and results on filtered How2 evaluation sets}
\vspace{-3pt}

\begin{table}[t]
  \centering
  \caption{Number of samples contained in each evaluation set filtered with various threshold $\alpha$.\vspace{5pt}}
  \begin{tabular}{c|cccccc}
    $\alpha$ & 0.5 & 0.6  & 0.7  & 0.8  & 0.9  & 1.0 (full) \\
    \hline
    \# samples & 110 & 441 & 921  & 1336  & 1666  & 2127
  \end{tabular}
  \label{table:filtered_num}
  \vspace{-7.5pt}
\end{table}

While we showed the effectiveness of our system in Section \ref{sec:result1}, we observed several evaluation examples with extraordinarily high scores only for the E2E SSum models compared with the cascade model.
We analyzed these samples and found that the training dataset contained samples with similar or perfectly matched summaries to the evaluation set.
For example, the summary of ID: 6D0hfwYIwm4 from the training set:

\begin{table}[h!]
  \vspace{-1pt}
  \hspace{0.5cm}
  \begin{tabular}{l}
    \textit{\small{Indian food is full of interesting spices and flavors.}} \\
    \textit{\small{Learn how to \textbf{serve} Indian potatoes, carrots \& peas ...}}
  \end{tabular}
  \vspace{-1pt}
\end{table}

\noindent{is very similar to that of ID: 2NdV94T-JNc from the evaluation set:}
\begin{table}[h!]
  \vspace{-1pt}
  \hspace{0.5cm}
  \begin{tabular}{l}
    \textit{\small{Indian food is full of interesting spices and flavors.}} \\
    \textit{\small{Learn how to \textbf{season} Indian potatoes, carrots \& peas ...}}
  \end{tabular}
  \vspace{-1pt}
\end{table}

\noindent{The audio data corresponding to such summaries are often different parts of a single video (e.g., the first and second parts of a video are in the training and evaluation sets, respectively.)}
This is possibly because the How2 dataset defines the ``description'' tags of YouTube as the summary, as mentioned in Section \ref{sec:dataset}, and the training and evaluation sets are partitioned randomly.
Under this condition, it is difficult to evaluate the generalization capabilities of the models. For example, a model overfitting the training data could achieve high scores on the test samples that overlap with the training set.

Therefore, we propose to filter the evaluation set to omit such samples by utilizing a quantitative criterion.
First, we computed the ROUGE-L scores between each summary in the evaluation set and the whole How2 dataset. 
We defined the highest score as the \textit{leakage score} for each summary in the evaluation set.
For example, the leakage score of ID: 2NdV94T-JNc is 0.95.
Then, we removed the samples whose leakage score was larger than a threshold $\alpha$ from the evaluation set.
Table \ref{table:filtered_num} shows the number of samples contained in each filtered evaluation set for different threshold values.

\begin{figure}[t]
  \hspace{-7pt}
  \centering
  \includegraphics[scale=0.345]{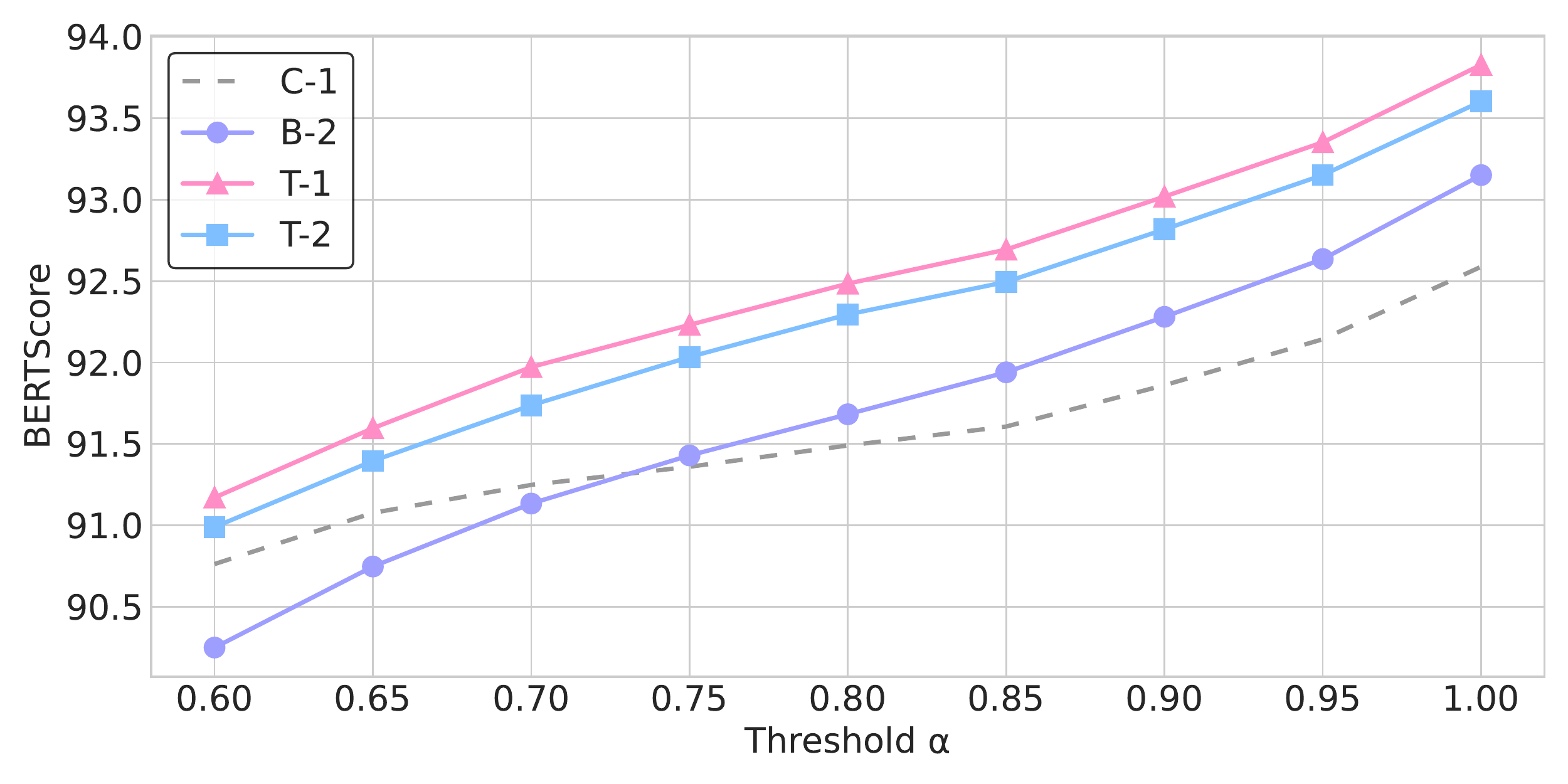}
  \caption{The METEOR (above) and BERTScore (below) scores of Models B-2, T-1, T-2, and C-1 changing the filtering threshold $\alpha$ from 0.6 to 1.0.}
  \label{fig:alpha}
  \vspace{-10pt}
\end{figure}

Figure \ref{fig:alpha} shows the METEOR and BERTScore scores of Models B-2, T-1, T-2, and C-1 with a different threshold $\alpha$.
While the scores for all models tend to be worse as $\alpha$ decreases, Model B-2 shows the clearest trend. 
This result suggests that the baseline E2E system overfits the training data and is vulnerable to unfamiliar samples.
Model C-1 shows robustness for unseen samples and the slowest score reductions, probably owing to the abundant linguistic knowledge of its TSum model.
For our proposed method, Models T-1 and T-2 achieved the highest accuracy in all conditions and seemed to have gained linguistic information from the external TSum corpus.

\begin{figure}[t]
  \centering
  \hspace{-7pt}
  \includegraphics[scale=0.345]{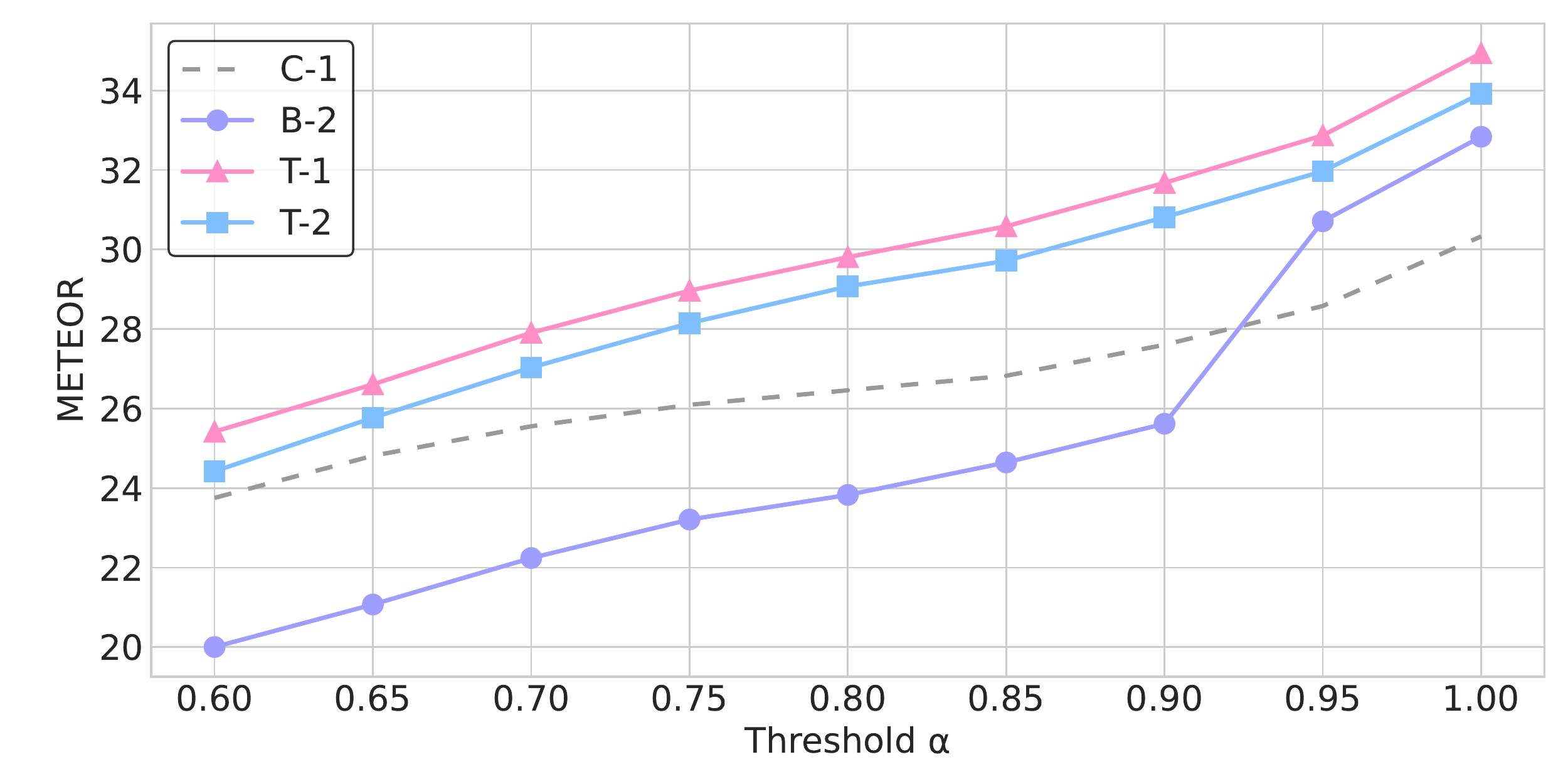}
  \vspace{-20pt}
\end{figure}

\vspace{-3pt}
\section{Conclusion and Future Work}
\vspace{-3pt}
In this paper, we studied the effectiveness of data augmentation methods for E2E SSum with external text summarization data on the How2 dataset. 
Compared with a strong baseline, the TTS- and phoneme-based data augmentation methods improved the METEOR score by 2.1 and 1.1 points, respectively. 
In addition, we also showed a clear trend of acquiring linguistic knowledge by utilizing TSum data in our E2E SSum model.

In this paper, we leverage only text summarization data as input for the E2E SSum models. 
As part of future research, we will attempt to leverage large external language models to feed the models with knowledge from diverse unpaired text data.

\vfill\pagebreak

{\small
  \bibliographystyle{IEEEbib}
  \bibliography{strings,refs}
}

\end{document}